# Medical idioms for clinical Bayesian network development

Evangelia Kyrimi[1,2], Mariana Raniere Neves[1], Scott McLachlan[1], Martin Neil[1], William Marsh[1], Norman Fenton[1]

[1] *Risk and Information Management Research Group, School of Electronic Engineering and Computer Science, Queen Mary University of London, London, UK, E1 4NS.*
[2] *Corresponding author e-mail address: e.kyrimi@qmul.ac.uk*

## Abstract

Bayesian Networks (BNs) are graphical probabilistic models that have proven popular in medical applications. While numerous medical BNs have been published, most are presented *fait accompli* without explanation of how the network structure was developed or justification of why it represents the correct structure for the given medical application. This means that the process of building medical BNs from experts is typically *ad hoc* and offers little opportunity for methodological improvement. This paper proposes generally applicable and reusable medical reasoning patterns to aid those developing medical BNs. The proposed method complements and extends the idiom-based approach introduced by Neil, Fenton, and Nielsen in 2000. We propose instances of their generic idioms that are specific to medical BNs. We refer to the proposed medical reasoning patterns as *medical idioms*. In addition, we extend the use of idioms to represent interventional and counterfactual reasoning. We believe that the proposed medical idioms are logical reasoning patterns that can be combined, reused and applied generically to help develop medical BNs. All proposed medical idioms have been illustrated using medical examples on coronary artery disease. The method has also been applied to other ongoing BNs being developed with medical experts. Finally, we show that applying the proposed medical idioms to published BN models results in models with a clearer structure.

## 1. Introduction

Each year many hundreds of novel Bayesian Networks (BNs) are presented in the literature, with medicine being a popular application domain [1], [2], [3], [4], [5]. Most published BNs are present as *faits accomplis*: with little explanation of how the network structure was developed and without justifying whether the structure is correct for a given scientific application. This problem is even more evident for medical BNs [5]. The process of building medical BNs, with medical experts, is typically *ad hoc* and offers little opportunity for repeatability and standardisation. However, a clear description of the modelling approach is necessary if clinical and patient communities are to adopt the methodology or resulting BN into healthcare practice.

Building a BN involves two main tasks [6]: (1) determine the BN structure; and (2) specify the parameters. Both the BN structure and parameters can be built: (a) by automated learning from data if sufficient data are available [7], [8], [9]; (b) "by-hand" using knowledge elicitation methods to capture domain expert knowledge and extract necessary information from the literature [1], [10], [11], [12]; or, (c) through a combination of both [13], [14]. In many medical problems it is not feasible or appropriate to use automated techniques and the structure and/or parameters of the BN must be elicited from experts. Much research has focused on knowledge elicitation methods and considers how to elicit parameters from experts [15], [16] [17], [18], [19], [20], or statistically estimate them [21], [22], [23]. However, as articulated convincingly in [24], less attention has been paid to the problem of eliciting the BN structure and, in particular, causal knowledge before considering any learning from data [25]. Hence, in summary, the literature lacks a systematic, repeatable method for building medical BNs from experts and literature.

There are impediments that make determining the *right* structure for a BN difficult, and several groups have proposed guidelines to help avoid spurious or unreliable models[26], [27]. Neil *et al*. found that by applying natural reasoning patterns, called *idioms*, BN development could be faster and deliver better-quality BNs. Idioms are abstract reasoning patterns from which specific cases, called *instances*, can be constructed. For example, instances of idioms have been used to develop BNs for assessing legal arguments [28], [29]. While Neil *et al's* generic idioms are capable of broad application to many subject domains, this paper proposes new types of idioms specifically to develop medical BNs. We refer to these as *medical idioms*, and we extend the use of idioms to represent additional types of reasoning relevant in medical decision-making, including: (1) interventional; and, (2) counterfactual reasoning.

This paper proposes a systematic and repeatable method for developing medical BNs that both complements and extends Neil *et al's* idiom-based approach. We believe the proposed medical idioms can provide a basis for standardised and consistent development of medical BNs. The remainder of this article is organised as follows: in Section 2 we provide an overview of the foundations of BNs and idioms. In Section 3 related work is presented. Our proposed medical idioms are explained in Section 4, and we include an example of how they are combined into a core BN model of coronary artery disease. Assessment of medical idioms against established BN models is presented in Section 5. All BN examples and models were implemented in AgenaRisk [30]. Our conclusions and recommendation for further work are presented in Section 6.

## 2. Background: Bayesian networks and idioms

### 2.1 Bayesian Networks

A BN is a directed acyclic graph with qualitative and quantitative parts [31]. The qualitative part is the graph comprised of nodes representing random variables (discrete or continuous) and directed edges representing causal or influential relationships. If a directed edge connects variables A and B, such as A → B, then A is called parent node or ancestor of B, and B is a child node or a successor of A. For instance, in Figure 1a smoking is a parent of lung cancer, indicating that smoking has a causal impact on lung cancer. The quantitative part of a BN comprises a set of conditional probability functions associated with each node – captured by a Node Probability Table (NPT) - to represent the conditional probability distribution of each node in the BN given its parents (Figure 1b).

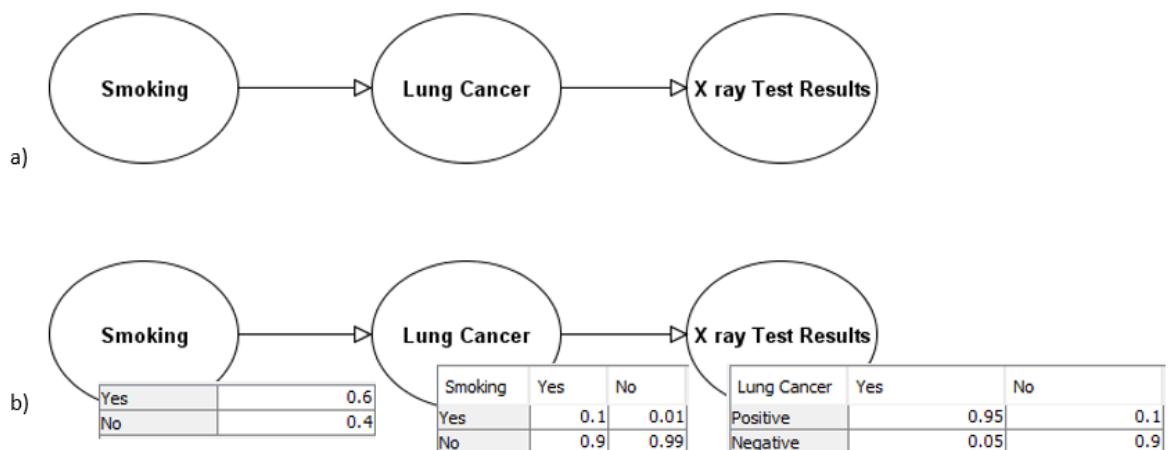

*Figure 1. (a) A three-node BN example. (b) A three-node BN example with Node Probability Tables shown*

Once values for all NPTs are provided the BN is fully parameterised and a variety of probabilistic reasoning processes can be performed. Bayesian probabilistic reasoning describes the process of updating our prior belief about an uncertain hypothesis in light of new evidence. Our initial belief is termed *prior probability*, while our updated belief is termed *posterior probability*.

Once evidence is entered in the BN, the probabilities of remaining unobserved variables are updated. There are two ways of reasoning when evidence is entered:

1. *Forward reasoning*: The reasoning process follows the direction of the arc. For instance, knowing that the patient is a smoker, increases the probability of lung cancer from 6.4% (Figure 2a) to 10% (Figure 2b). Forward reasoning is described as *causal reasoning* when the BN structure represents true causal relationships rather than simple associations, which is not an absolute requirement when we reason from evidence.
2. *Backward reasoning*: The reasoning process is counter to the direction of the arc. For instance, knowing that the patient's X-ray is positive increases the probability of lung cancer (Figure 2c). Background reasoning is also described as *diagnostic reasoning* when the BN structure represents true causal relationships.

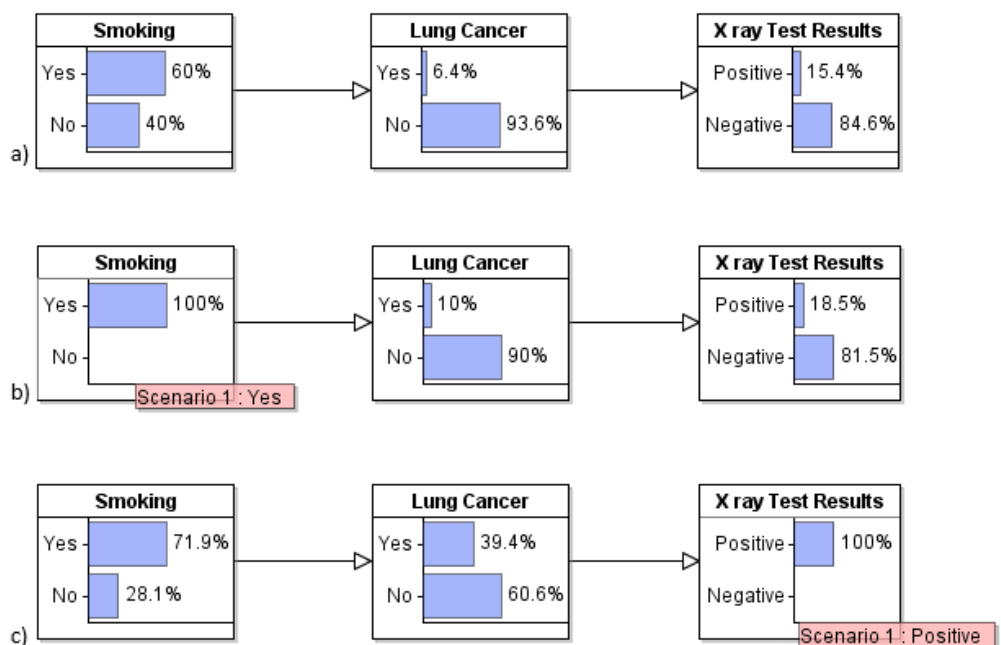

*Figure 2. (a) BN example with marginal prior probabilities. (b) BN example, where causal reasoning is performed (Scenario 1 appears when evidence is entered in AgenaRisk). (c) BN example, where diagnostic reasoning is performed*

A particular case of diagnostic reasoning is a phenomenon called *explaining away* [32], or *discounting* [33]. Explaining away occurs when a child node has more than one independent parent node. If the child node, or one of its ancestors, is observed, the probability of each of the parent nodes changes (diagnostic reasoning). But if one of the parents is known, then the probabilities of the other parents change. This is because the change on the child node can be explained by changes in either of the parent nodes, hence making the parent nodes dependant.

BNs can also be used to answer hypothetical questions such as *what will happen if an intervention is made*. Reasoning about interventions can only be performed when relationships among the variables are causal because an intervention is an exogenous action that fixes the state of the variable we have intervened upon, making it independent of its causes [34], [31], [35]. Contrary to reasoning about evidence, interventional reasoning does not allow diagnostic reasoning from the intervened variable [36]. For instance, when we observe a high body temperature on the thermometer, we can argue that we have fever. However, if we arbitrarily start rubbing the thermometer to reach a specific temperature, it can no longer be argued that fever is present. According to Pearl [31], an externally imposed intervention is presented using the *do operator*, as in the previous example: the probability of having fever given that we intervened by rubbing the thermometer is presented as $P(Fever = Yes|do(Thermometer = High\ Body\ Temperature))$. The process of making the intervened variable independent of its causes by removing all the edges pointing towards that variable is known as *graph surgery*.

BNs can also be used to answer counterfactual questions such as *what would have happened if events other than those observed had happened*. In BNs, counterfactual reasoning combines both evidence and interventions. Using the process of *twin-networks* proposed by Pearl [31], the actual world is modelled based on evidence while the hypothetical counterfactual world is altered using interventions. In a twin network both networks have identical structures except for the arrows towards the variable that we intervene upon, which are missing in the hypothetical world. The posterior probabilities of the variables, which remain the same in both worlds, are called *background variables,* and are shared between the two networks. The main difference between counterfactuals and interventions is that for the former we know the values that some or all the variables had in the actual world. By contrast, when we intervene, we are unaware of the values of the predecessor variables in the network.

## 2.2 Idioms

Neil et al. [26] introduced an idiom-based approach to describe how elicited variables should be connected. In their book [37], Fenton and Neil present four idioms that cover a wide range of modelling tasks:

1. *Cause-consequence idiom:* Models uncertainty of a causal process with observable consequences. The cause-consequence idiom has a chronological order: the consequence follows the cause and not the other way around. For instance, as shown in Figure 3, a car accident causes head injury.

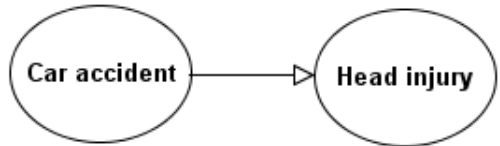

*Figure 3. A medical example of the cause-consequence idiom proposed by Fenton and Neil (2018)*

2. *Measurement idiom:* Models uncertainty surrounding the accuracy of any type of measurement. The causal directions here can be interpreted in a straightforward way. The actual value of the attribute must exist before it is measured (by a person or machine). The measurement has a known accuracy. The higher the accuracy, the closer the assessed value is expected to be to the actual value of the attribute. For example, as shown in Figure 4, the result of an X-ray test is a measurement for assessing internal chest bleeding. Although a positive or negative result is strongly indicative of whether there is internal chest bleeding or not, there are inevitably false positives and false negatives. In Figure 4a the full version of the measurement idiom is presented, where the X-ray accuracy is either standard (having 1% false positive and 5% false negative rate), or perfect. In Figure 4b, only a standard X-ray test is assumed. Diagnostic reasoning is performed. The latent or difficult to observe/measure attribute becomes a special case of the measurement idiom.

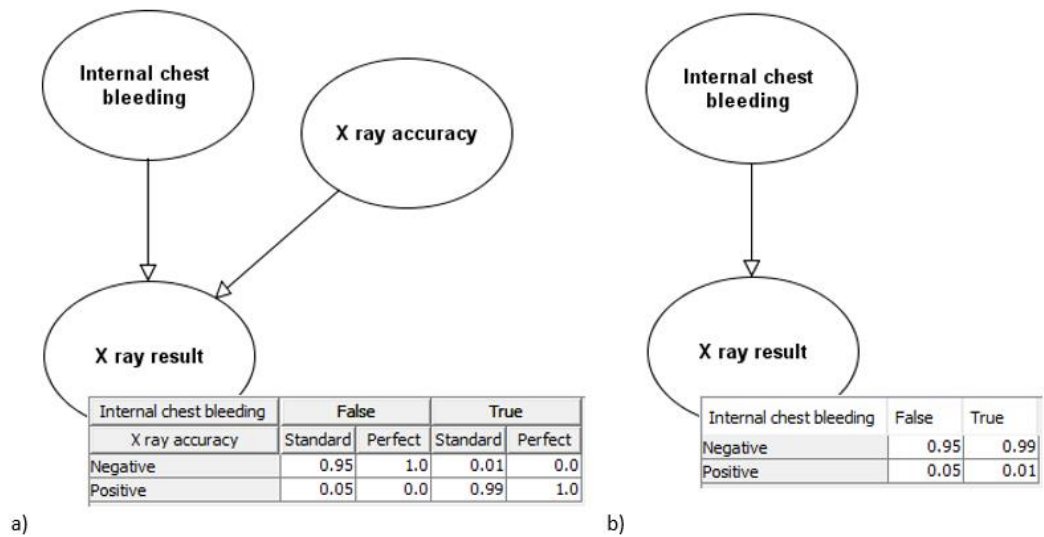

*Figure 4. (a) A medical example of the standard measurement idiom proposed by Fenton and Neil (2018). (b) A medical example of the implicit measurement idiom proposed by Fenton and Neil (2018)*

3. *Definition/synthesis idiom:* Models synthesis or combination of many nodes into one synthetic to cover the following cases: (1) defining a synthetic node in terms of its parents (Figure 5a); (2) combining variables using a hierarchical structure (Figure 5b); or, (3) combining different nodes together to reduce the parameters in the NPTs (Figure 5c).

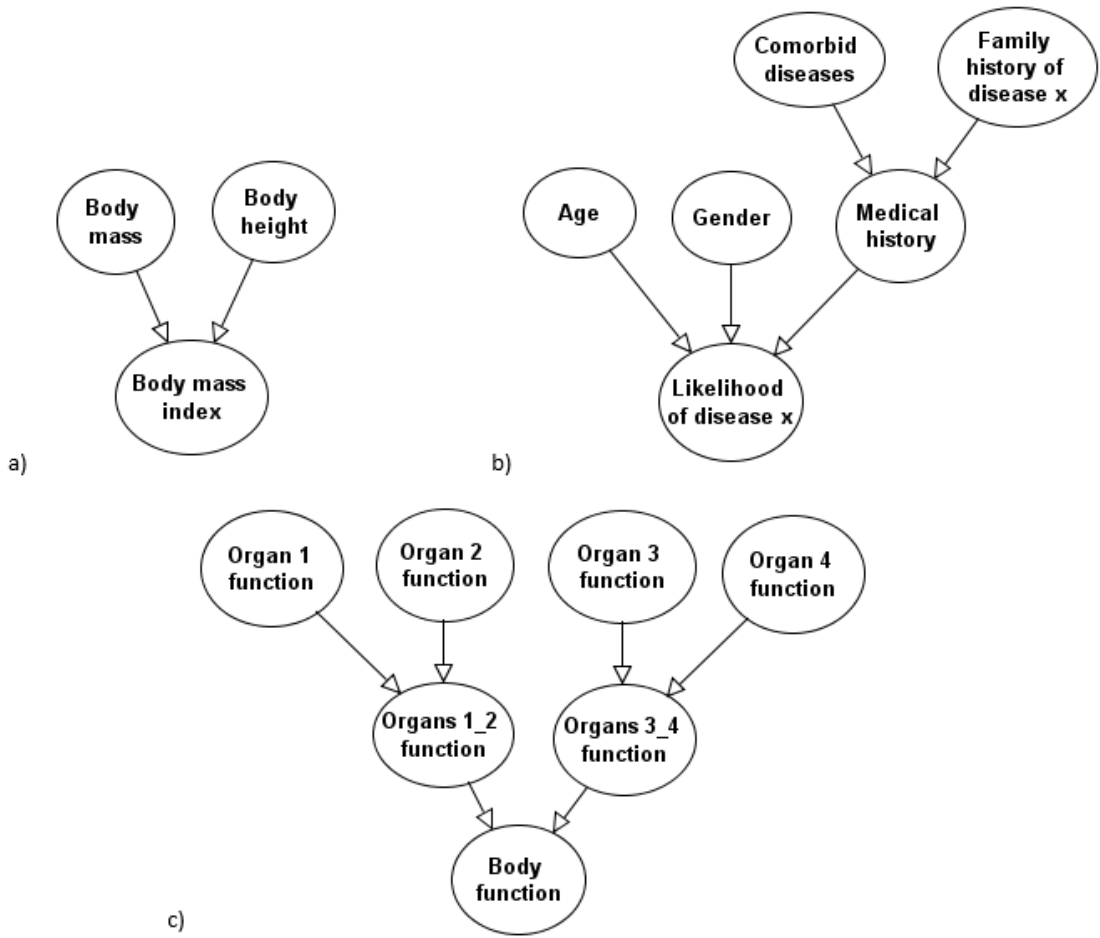

*Figure 5. (a) A medical example of the definitional/ synthesis idiom for case 1; definitional relationship between variables, proposed by Fenton and Neil (2018). (b) A medical example of the definitional/ synthesis idiom for case 2; hierarchical definitions, proposed by Fenton and Neil (2018). (c) A medical example of the definitional/ synthesis idiom for case 3; reduce effects of combinatorial explosion, proposed by Fenton and Neil (2018)*

4. *Induction idiom:* Models uncertainty related to inductive reasoning based on populations of similar or exchangeable members. The induction idiom is a general model for any type of statistical inference represented using a BN. The main difference is that an unknown or partially known population parameter is learned from known data (Figure 6).

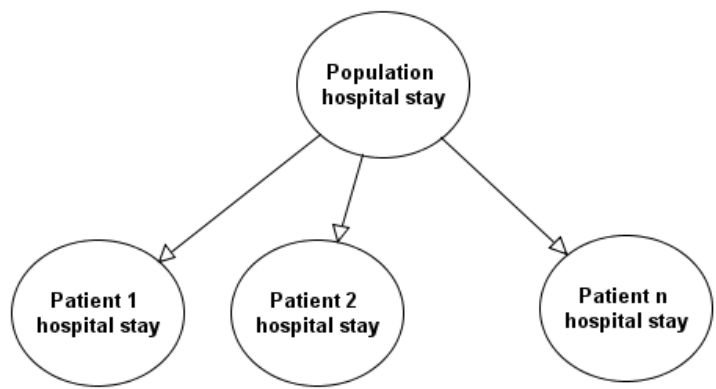

*Figure 6. A medical example of the induction idiom proposed by Fenton and Neil (2018)*

## 3. Related work

Several knowledge engineering methods have been proposed to support development of complex BN structures. For example, Laskey and Mahoney [38], [39], [40] recognised that BN construction requires a method for specifying meaningful building blocks which they called *network fragments*. A network fragment consists of a set of related random variables together with knowledge about how they are related. Ideally, fragments should make sense to the domain expert who must supply some underlying motive or reason for their existence. Koller and Pfeffer [41] proposed an abstract approach to support the construction of large BNs, known as *Object-Oriented BNs* (OOBNs). An OOBN is made up of classes that contain both ordinary BN nodes as well as abstract objects, representing instances of other classes. A class can be considered as a network fragment. OOBNs are particularly useful for complex models that contain repeated fragments, where objects can be reused to decrease the modelling effort. Network fragments and OOBNs are both useful when organising and decomposing a large, complex BN, but they do not explain the connection and reasoning among variables elicited from domain experts.

More guidance for developing a BN is given from the following types of dependency connection, known also as *d-connection*:

1. *Serial connection*: C is conditionally dependent on B, which is conditionally dependent on A. A and C become conditionally independent, also known as *d-separated*, given that B is known (Figure 7a).
2. *Diverging connection*: A and C are conditionally dependent on B. A and C become conditionally independent given that B is known (Figure 7b).
3. *Converging connection*: B is conditionally dependent on A and C. A and C are conditionally dependent only when B or any of its descendants, if available, are known (Figure 7c).

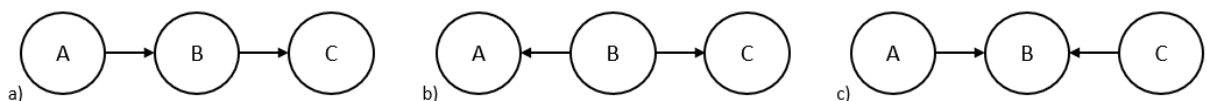

*Figure 7. (a) Serial connection. (b) Diverging connection. (c) Converging connection*

These three conditional dependencies can aid in connecting a small number of variables and verifying the reasoning process between them. However, in practice, experts do not think in terms of conditional dependencies. Medical BNs are complex and are composed from many variables. Thus, using the proposed dependency connections as the only guidance for developing medical BNs can be problematic.

Few semantic approaches have also been proposed for assisting the development of BNs. Kjaerulff and Madsen [25] propose an overall causal BN structure using abstract variables classes. Other researchers have explained the BN structure using abstract templates, where relationships between elicited variables were implicitly described. For instance, Helsper and van der Gaag [42] propose a methodology for building a BN for the domain of oesophageal cancer from ontologies. Derivation of the structure from the proposed ontology is conducted by addressing semantically meaningful units of knowledge. Seixas et al. proposed a three-level diagnostic BN structure [43]. Shwe *et al.* proposed a *Quick Medical Reference* (QMR) network built on statistical and expert knowledge [44]. The QMR belief network is a two-level graphical model, where the top level contains binary nodes representing diseases and the bottom level contains binary nodes representing findings. There are several conditional independence assumptions that restrict the QMR network structure. First, the diseases are assumed to be marginally independent, thus it does not entertain established comorbidities routinely observed in clinical practice. Second, findings are assumed to be conditionally independent given the known diseases' states. Third, and finally, findings are all modelled as manifestations of the disease, an assumption that is incorrect when modelling historical findings. In a later version of QMR network proposed by Pradhan et al. [21], background variables representing predisposing factors were included. This resulted into a three-level BN structure. Velikova et al. [45] present a template on how to model temporal BNs where the relationships between a few key medical variables were described. Still other have instead focused on proposing a pattern for modelling specific parts of a medical BN, such as the coexistence of two or more comorbid conditions [46], [47]. Luciani and Stefanini [48], [49] propose an automated interview for eliciting specific medical variables from experts and a way to connect them. The described templates and semantics are beneficial under certain circumstances; however, they are either too narrow, targeting only specific variable types, or they offer an overall structure without some underlying method or an explanation of the reasoning patterns to encourage reusability.

## 4. Medical idioms

As illustrated in Figure 8 and established from literature on the classical approaches to diagnosis and treatment, while there is much information and many tests that can help clarify the medical condition [50], [51], it remains difficult for humans alone to detect, comprehend and correctly identify the presence of the underline cause [52], [53]. It is in the feedback loops, presented in Figure 8, where conditions are diagnosed and treatments selected that many believe Artificial Intelligence (AI), of which BNs are one type, can help [54], [55], [56]. Further, it is as a novel component of the resource-intensive and multidisciplinary process of developing that AI where we position our medical idioms as one methodology capable of helping bridge the barriers between medical knowledge and decision science, with the effect of simplifying the elicitation task.

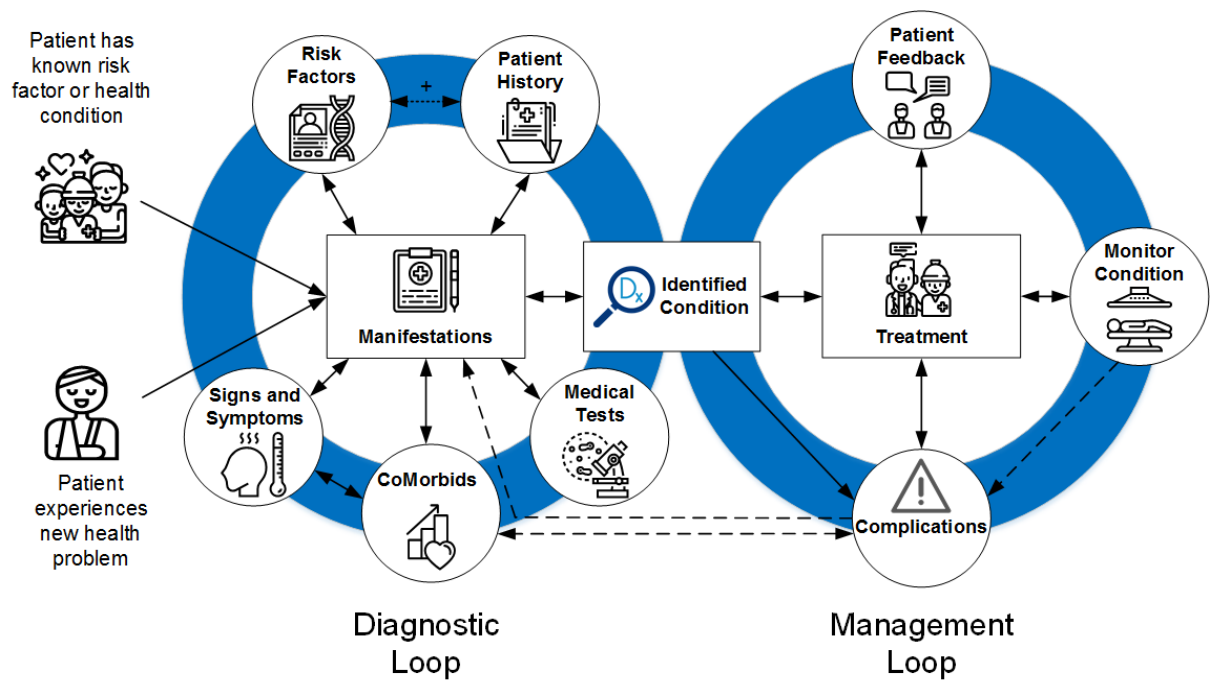

*Figure 8: The Classical diagnostic and management loops*

A medical BN should capture all relevant clinical variables and follow the causal mechanisms of the medical problem and care as described in Figure 8. When building a medical BN using knowledge elicitation methods, the model expert follows logical causal patterns to connect elicited variables. Idioms represent natural and reusable reasoning patterns that follows human reasoning processes. Consequently, idioms form the basis of our methodology for developing medical BNs. However, each generic idiom presented in Section 2.2 covers many medical scenarios that should be differentiated. Therefore, we propose instances of the generic idioms that represent specific medical patterns. The BN structure, as well as the medical idioms, are highly related to the type of reasoning that is taking place. For that reason, we extend the use of idioms beyond only reasoning about evidence. We propose medical idioms that represent additional types of reasoning, such as interventional and counterfactual reasoning - analyses which are not uncommon in medical applications.

Idioms act as a library of patterns for BN development. The model expert connects elicited knowledge with idioms and reuses the most appropriate. Consequently, for developing medical BNs the model expert must first understand the role each elicited variable plays as a component of clinical reasoning, and then select the medical idiom best describing the related reasoning process. So, before developing the BN structure using idioms, elicited variables should be classified, based on their role in clinical

reasoning as illustrated in the classical diagnostic and management feedback loops described in Figure 8:

1. **Conditions (C)** indicate states of health, whether normal or abnormal. A condition might be further distinguished as a disease, a disorder or a syndrome. However, we use the term condition as a broad-neutral term.
2. **Manifestations (M)** are observable consequences of a condition. They are regarded as part of the condition's semiotics and are classified according to the condition that triggers them. Manifestations can be further grouped into the following categories:
    a. **Symptom (Sy)** is a subjective feeling or departure from normal function which is apparent only to a patient (e.g fatigue)
    b. **Sign (Si)** is an objective indication of the condition that can be observed by clinicians.
    c. **Medical Test (MT)** is a procedure performed to diagnose or monitor a medical condition and/or a sign.
3. **Risk Factors (RF)** are observable attributes, characteristics or exposures of an individual that increase his likelihood of developing a medical condition or a manifestation of it. Risk factors are regarded as part of condition's aetiology and are classified according to the condition that they influence.
4. **Pathogenic Mechanisms (PM)** are unobservable mechanisms that describe the pathogenesis/ development of a condition. They are also part of a condition's aetiology as they explain the impact of some risk factors on the condition. Pathogenic mechanisms can be distinguished from conditions as they normally cannot explain manifestations [48]. However, as more diagnostic tools become available, the boundary between conditions and pathogenic mechanisms becomes weaker.
5. **Treatment (T)** represents the clinical management to treat or cure a condition, described as healthcare.
6. **Comorbidities (CC)** are medical conditions that exist simultaneously in the same patient [57]. The origin of a comorbidity may lay in the anatomical proximity of the diseased organs, the singular pathogenic mechanism of the number of conditions or the cause-effect relationship between conditions. Comorbidities play a crucial role in diagnosis and treatment recommendations [46], [47]. Two comorbid conditions can be entirely unrelated. On the other hand, when two comorbid conditions are related, the second or subsequent condition can cause, be caused, or be otherwise related to the first condition.
7. **Complications (Cm)** are unfavourable elevations or consequences of a condition or a treatment. The condition itself can exacerbate, manifesting more severe symptomology or even causing another condition, a comorbidity. A treatment can also result in adverse events. A complication is so named because it *complicates the situation*. Thus, accounting for it is an important step for prognosis or treatment selection.

The above classifications represent the information that clinicians normally use for describing a condition. Thus, these are variables that are usually present in a medical BN. We do not claim that a variable necessarily has a unique classification. The role of a variable in the clinical reasoning might vary in different medical contexts. For example, clinicians in a cardiac clinic may see obesity as a risk factor, while in another clinic obesity might be the condition in question. However, when developing a BN for a specific medical problem, a unique variable classification should be feasible, although there might still be cases where a variable has more than one classification. The multipurpose of a variable may be due to different variables that trigger, or are triggered by, it. For instance, a long bone fracture can be a sign of limb injury but also a risk factor for bleeding. This is not an issue when using medical idioms as overlap among idioms is allowed.

After classifying each elicited variable, the model expert should consider how the classified variables relate to one another. This should lead to subset of variables grouped together. Each subset should then be examined in terms of the flowchart shown in Figure 9 to determine which idiom is best represented. In the remainder of this section the proposed medical idioms are defined in detail.

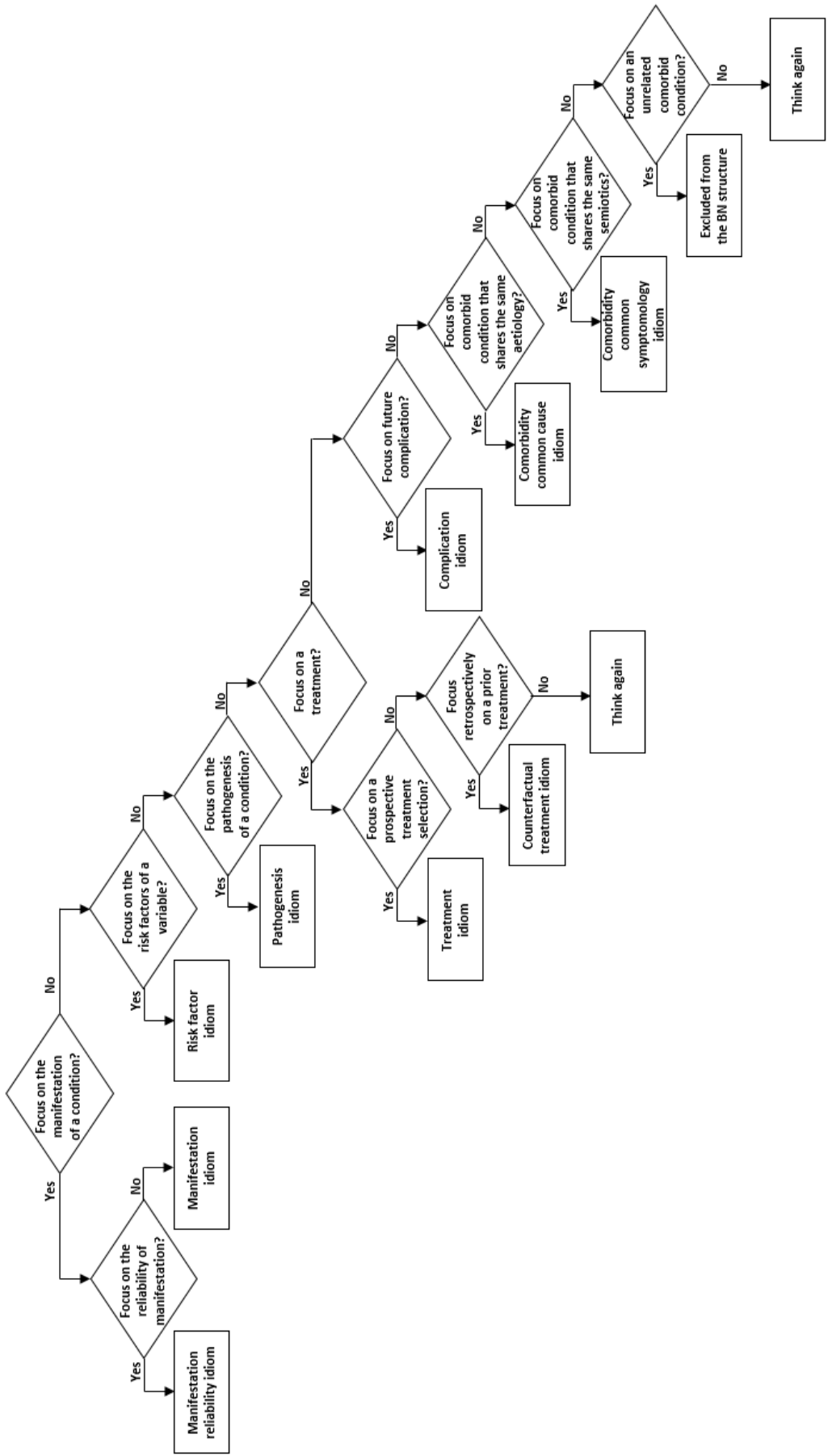

*Figure 9. Flowchart for selecting the right medical idiom*

## 4.1 Manifestation idiom

The manifestation idiom models the uncertain causal relationship between a condition and related manifestation variables. It can be considered an instantiation of the *measurement idiom*, where diagnostic inference is normally performed. As mentioned in the *measurement idiom* presented in Section 2.2, the condition should exist before its effects are manifested. Figure 10 illustrates an example of the examination idiom, where coronary artery disease is the unknown condition clinicians are seeking to diagnose or rule out using manifestation variables including medical test results and the patient's signs and symptoms.

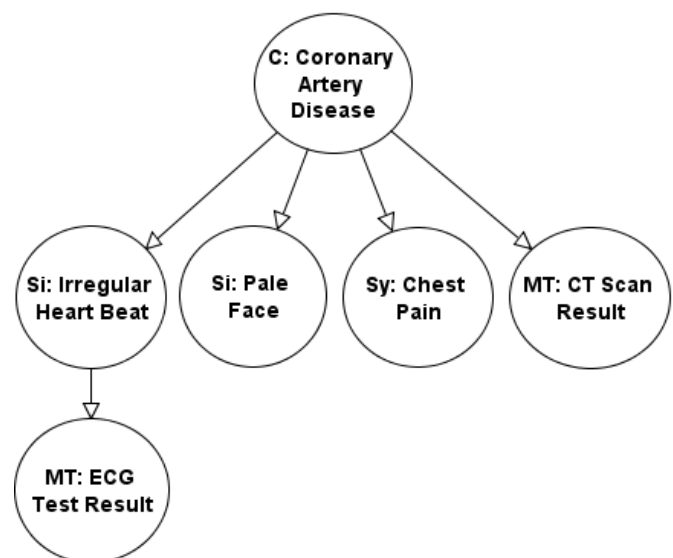

*Figure 10. An example of the manifestation idiom*

The inaccuracy of each manifestation variable, such as what an observed manifestation variable tells us about the condition, is captured as false positive and false negative rates. Another source of inaccuracy is the human actors' reliability in providing and interpreting the examination variables. This leads to the following extension of the manifestation idiom.

## 4.2 Manifestation reliability idiom

The manifestation reliability idiom models reliability of manifestation variables observed or interpreted by human actors. It makes explicit the notion that evidence provided or interpreted by human actors can be fallible. The manifestation reliability idiom shown in Figure 11 is an instance of the *measurement idiom*. The reliability of a reported symptom represents certainty on the degree of truthfulness and accuracy the observer uses to weight the patient's description of their own condition. The reliability of a symptom might be defined by many factors, such as the patient's objectivity and veracity (Figure 12a). This version of the idiom can also be represented using an object-oriented structure as shown in Figure 12b. The proposed decomposition of reliability is simply an instance of the *definition/synthesis idiom*. Patient's objectivity and veracity can be modelled either as Boolean

(Yes/No) or ranked nodes (e.g. underestimate, correct, overestimate). Ranked scales are particularly useful for chronic conditions. For example, suffering from a chronic condition can also bring about concomitant physiological and psychological disorders such as depression. A patient with depression may overestimate some of their symptoms, such as fatigue. Alternately, a patient suffering from a chronic condition may become conditioned to some symptoms. For instance, someone suffering with arthritis can become used to pain. This can result in underestimation of recurring pain. When a clinician treats that patient for many years, they may recognise whether a patient's self-assessment needs to be recalibrated for under- or overestimation. The reliability of a sign represents a clinician's certainty of observation. If we do not assume the clinician's reliability to be perfect, then the examination reliability idiom can be extended in a similar way to symptom reliability as shown in Figure 12. Reliability of a medical test result represents the clinician's potentially unreliable and subjective judgment when interpreting, whether the test result was positive or negative.

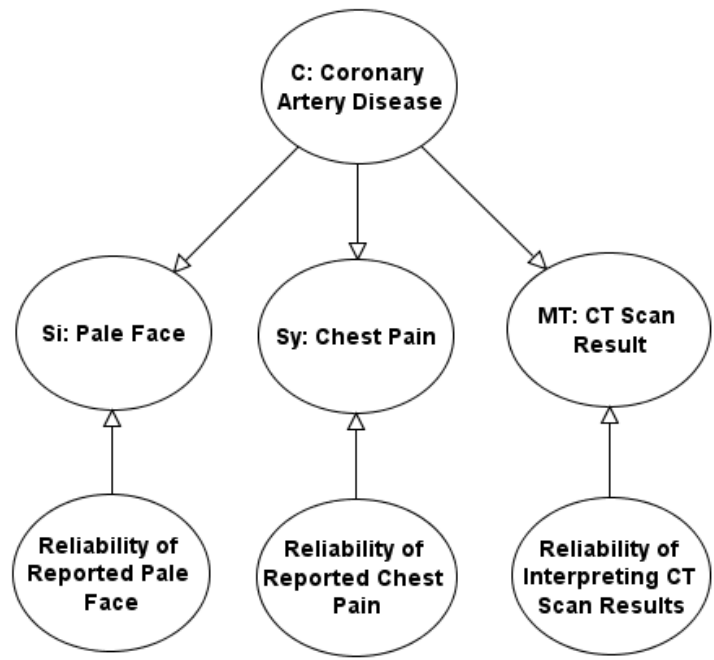

*Figure 11. An example of the manifestation reliability idiom*

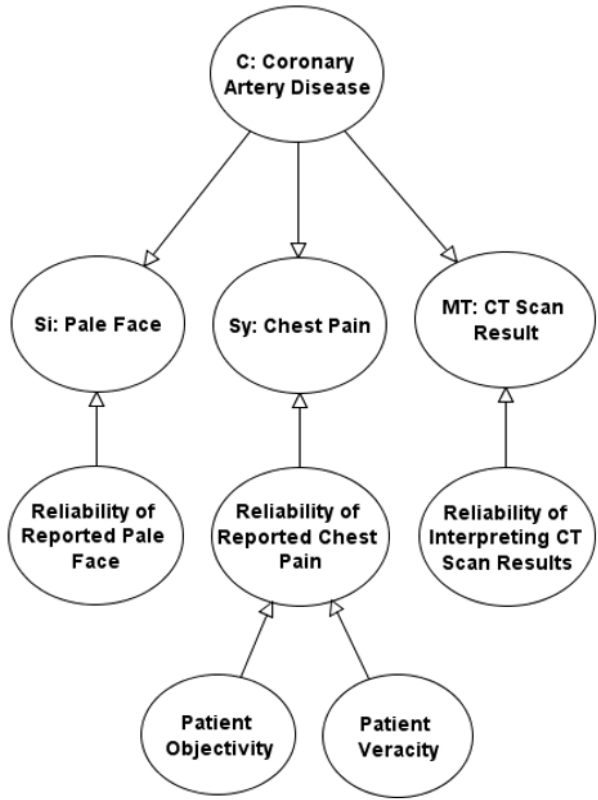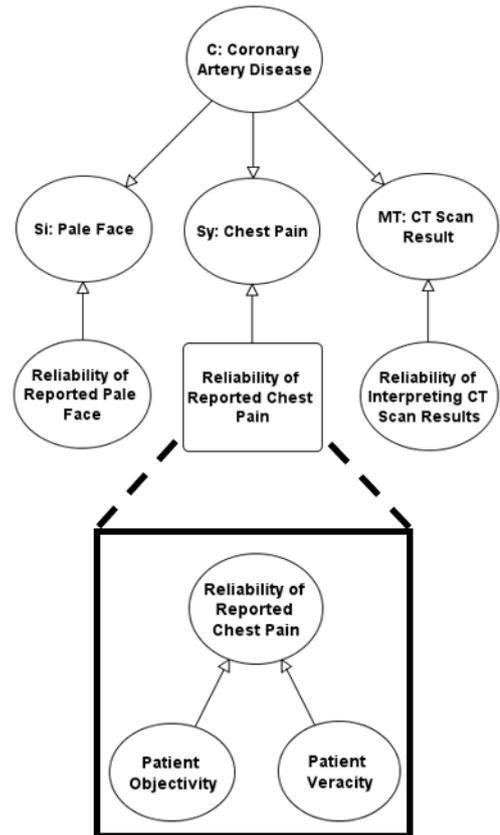

*Figure 12.(a) Symptom reliability idiom. (b) Symptom reliability idiom using an object-oriented structure*

The more reliable any manifestation variable provided by human actors is, the closer we can consider it to the real condition. Figure 13a shows an example of the manifestation reliability idiom where no evidence on reliability of the reported examination variables is entered. When we know that the reported examination variables are reliable, certainty for the presence of the condition increases (Figure 13b). However, when the reported manifestation variables are not reliable their impact on the condition diminishes (Figure 13c). When more than one sign, symptom, or diagnostic test is available, common reliability variables might be used to influence all the signs, symptoms, and diagnostic tests separately (Figure 14). This is because a patient's reliability on the presence of one symptom may affect confidence in the reliability of another, and equally, a clinician's confirmation bias about one sign or judgement regarding a test result may affect their reliability about another.

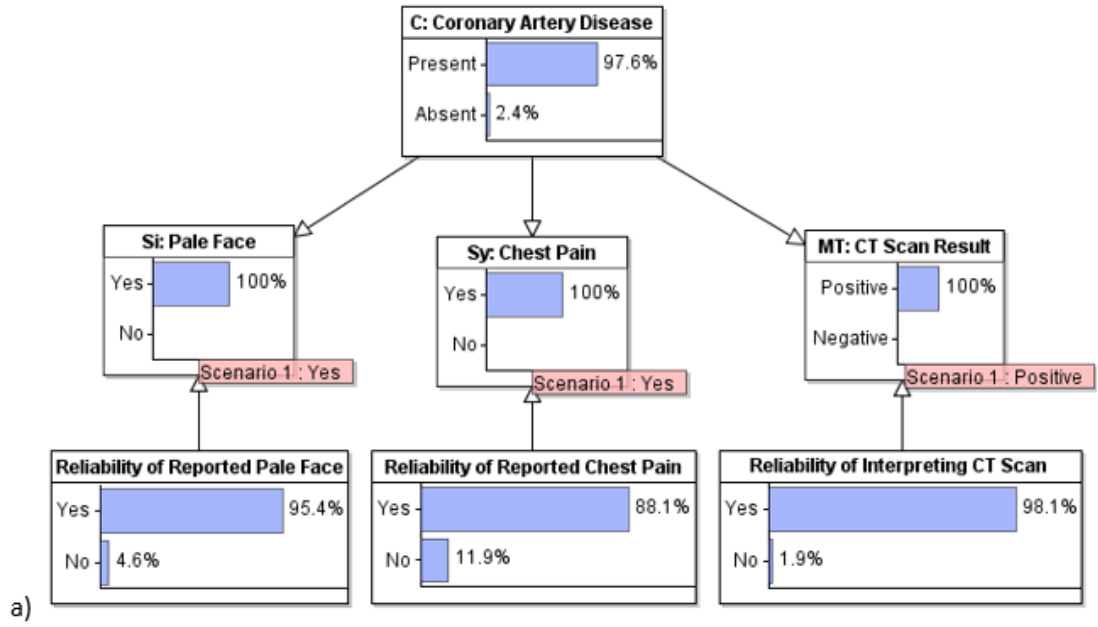

a)

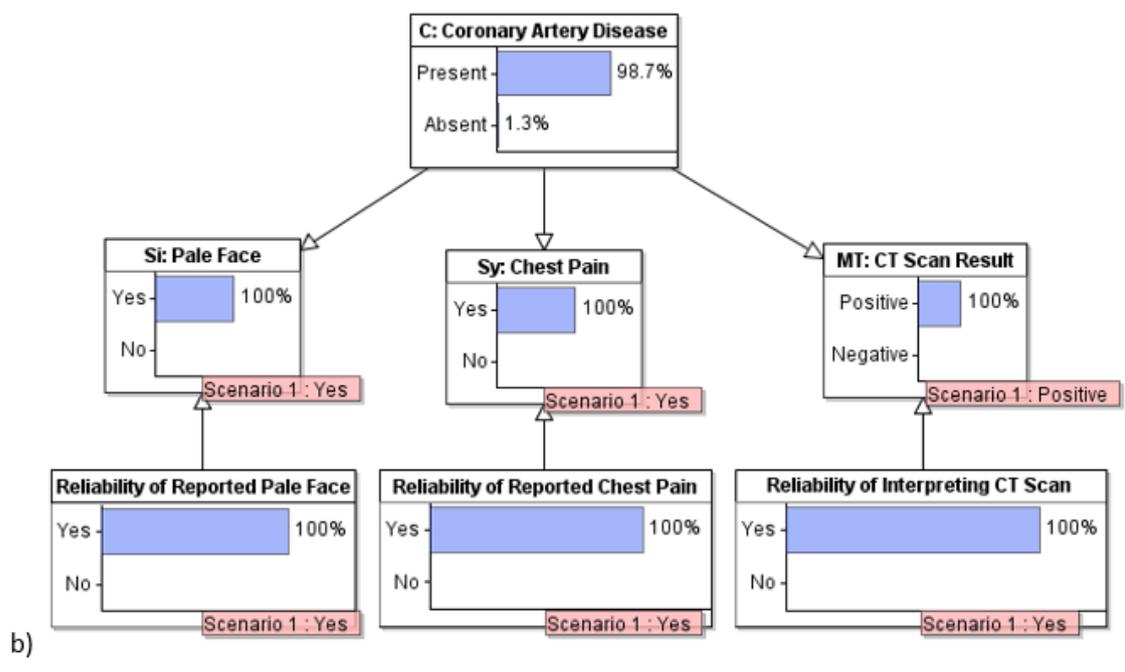

b)

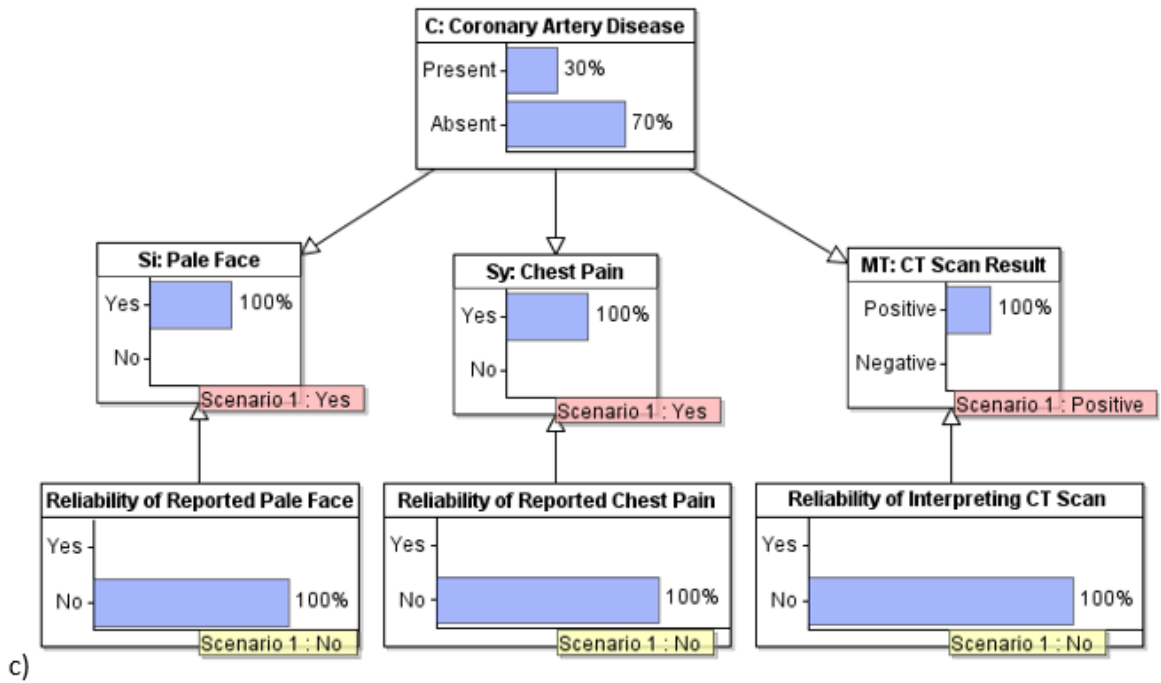

c)

*Figure 13. (a) An example of the manifestation reliability idiom when no evidence about the reliability is available. (b) An example of the manifestation reliability idiom when reliable evidence is available. (c) An example of the manifestation reliability idiom when non-reliable evidence is available.*

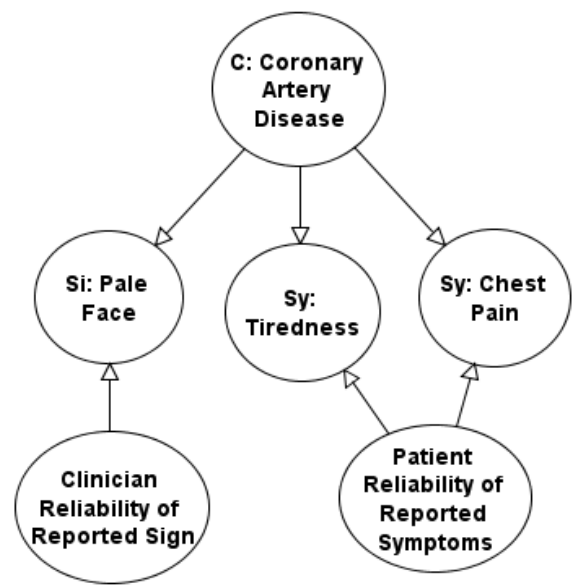

*Figure 14. Example of the manifestation reliability idiom with common reliability variables*

## 4.3 Risk factor idiom

The risk factor idiom models the uncertain relationships between an observable risk factor and the variables it affects (Figure 15). Risk factors occur before, or simultaneously with, their consequences. A risk factor can be an observable attribute or pre-existing comorbid condition that increases someone's likelihood of developing a condition, such as having a family history of cardiac problems increases the likelihood of developing a coronary artery disease. A risk factor can also be the reason why a manifestation variable has arisen or become exaggerated and can also be an inhibiting or supporting factor for choosing a treatment.

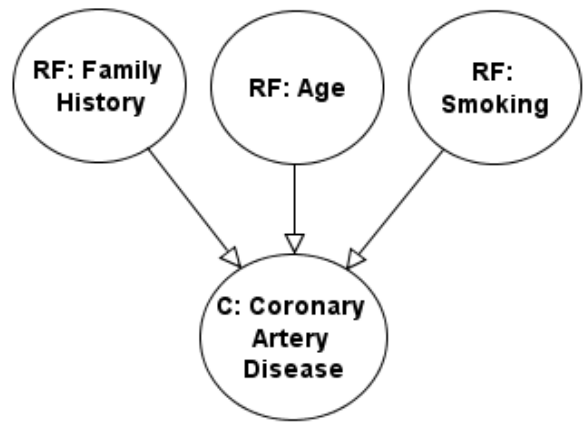

*Figure 15. An example of the risk factor idiom*

### 4.4 Pathogenesis idiom

The pathogenesis idiom models hidden underlying relationships between a risk factor and the condition. Sometimes a risk factor influences directly upon an individual's chances of developing the condition (modelled using the risk factor idiom). However, more often risk factors affect the condition indirectly through some unobserved pathogenic mechanism. For instance, obesity and diabetes are risk factors for developing coronary artery disease, however they do not cause the condition directly. Rather, they do so indirectly through the pathogenic mechanism of causing plaque to build up in arteries (Figure 16). As blocked arteries is an unobserved pathogenic mechanism, it cannot be considered a risk factor for coronary artery disease. Hence, the pathogenesis idiom is more appropriate than the risk factor idiom.

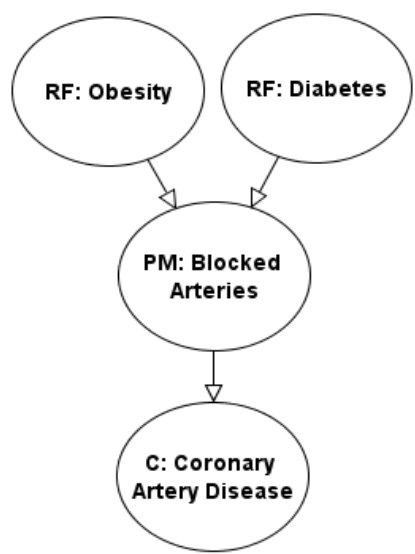

*Figure 16. An example of the pathogenesis idiom*

## 4.5 Comorbidity common cause idiom

The comorbidity common cause idiom models uncertain relationships between two conditions that share the same causes (Figure 17a). In this idiom one condition should always be a comorbid condition, while the other can be any condition group; a comorbid condition, the condition in question or a pathogenic mechanism. Based on the *cause-consequence idiom*, observing a cause, such as a risk factor, affects all its consequences (Figure 17b). The diverging connection between the two conditions makes them independent when their common cause is observed (Figure 17c). In case the common cause is not observed, then the two conditions become dependent (Figure 17d). The proposed comorbidity common cause idiom is similar to the *heterogeneity model* proposed by Valderas et al. [46].

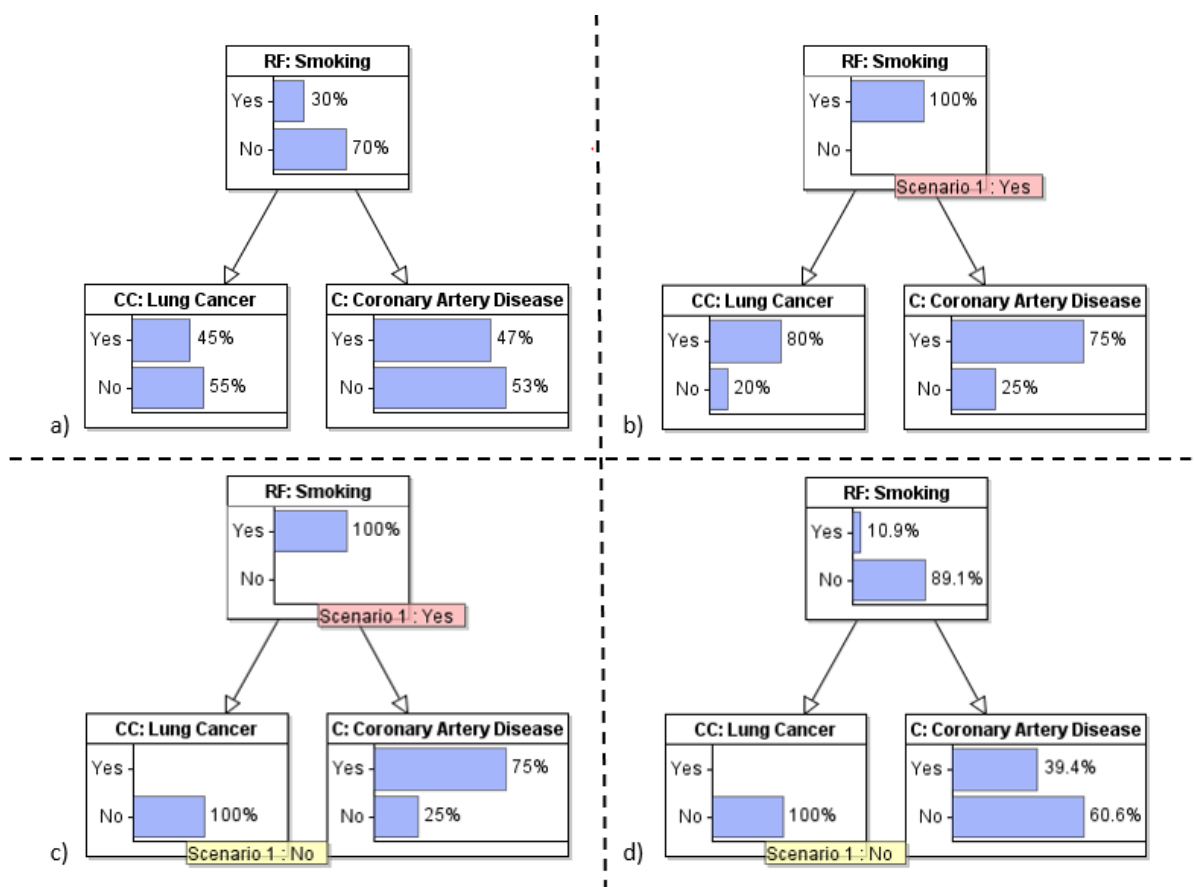

*Figure 17. (a) An example of the comorbidity common cause idiom. (b) An example of the comorbidity common cause idiom, where the cause is observed. (c) An example of the comorbidity common cause idiom, where the cause and the comorbid condition are observed. (d) An example of the comorbidity common cause idiom, where only the comorbid condition is observed*

## 4.6 Comorbidity common symptomology idiom

The comorbidity common symptomology idiom models uncertain relationships between two conditions that share the same consequences (Figure 18a). Like the comorbidity common cause idiom, one condition should always be a comorbid condition while the other can be any condition group: a comorbid condition; the condition in question; or, a pathogenic mechanism. When observing the shared consequences, such as a manifestation variable, both conditions are updated (Figure 18b). The converging connection between the two conditions makes them dependent when their common consequences are observed. As we can see in Figure 18c, knowing that the patient also has a lung cancer reduces the likelihood of a cardiac problem. In other words, having lung cancer is enough to explain the presence of chest pain. This is known as *explaining away* [58] or *discounting* [33]. In cases where the consequences are not evidence variables, the two conditions become independent (Figure 18d).

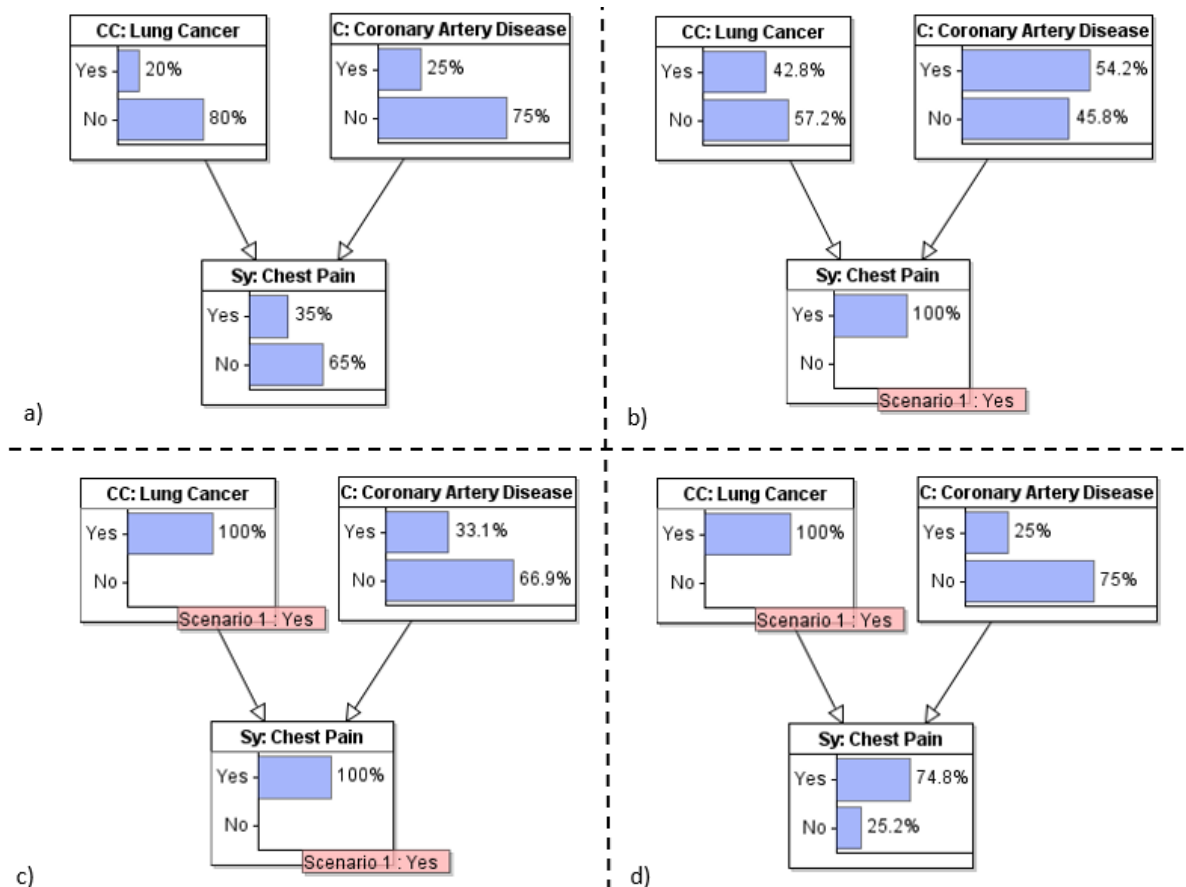

*Figure 18. (a) An example of the comorbidity common symptomology idiom. (b) An example of the comorbidity common symptomology idiom, where the consequence is observed. (c) An example of the comorbidity common symptomology idiom, where the consequence and the comorbid condition are observed. (d) An example of the comorbidity common symptomology idiom, where only the comorbid condition is observed*

## 4.7 Complication idiom

The complication idiom models the uncertain process between a condition or treatment and its unfavourable consequence. Timewise, the resulting consequence is a late effect. This idiom can be considered an instance of the cause-consequence idiom. An example of the complication idiom is shown in Figure 19. Heart attack can be an unfavourable future consequence of a coronary artery disease.

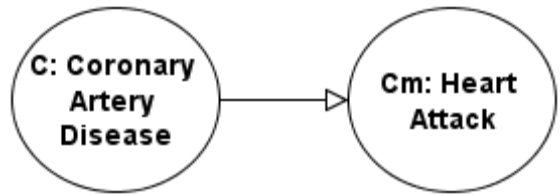

*Figure 19. An example of the complication idiom*

## 4.8 Treatment idiom

The treatment idiom models the uncertain process of condition management. A treatment is an important part of a BN as it describes a clinical decision. Figure 20a shows that suffering from coronary artery disease influences clinicians' decisions with respect to prescribing medication that reduces a patient's chances of having a heart attack. Given the medical context, a treatment can be affected by other comorbid conditions, risk factors, or previous treatment strategies. The treatment idiom is highly related to the type of reasoning process that takes place. For instance, if reasoning from evidence is performed and treatment is the hypothesis, then a structure such as the one shown in Figure 20a is adequate. However, if we want to capture the effect of a hypothetical treatment, then interventional reasoning should be performed, and a model with causal relationships is needed. Thus, following the graph surgery proposed by Pearl, the variable on which we intervene should be made independent of its causes. As a result, all the arcs pointing towards the intervened variable are removed, as shown in Figure 20b, and the effect of the hypothetical treatment $T = t$ on heart attack is estimated as $P(Cm = cm|do(t))$.

Typically, randomised control experiments [59] are used to learn cause-effect relationships from experimental data The randomisation process samples data by avoiding possible selection bias and confounding. In observational studies a subject that receives a treatment may do so because it has a more severe condition than those who remain untreated. Thus, any statistical association derived from observational data runs the risk of confounding the beneficial effect of the treatment and the underlying greater risk in those subjects who received the treatment. To arrive at an unconfounding treatment effect we must supress any confounder that influences both the treatment and the variable the treatment affects. Based on the example shown in Figure 20a, the variables *T* and *Cm* are unconfounded if and only if the following holds: $P(Cm = cm|do(t)) = P(cm|t)$ for all values $T = t$ and $Cm = cm$. This equality states that *T* and *Cm* are not confounded when the association observed in the data between the two variables is the same as the association that would have been measured in a randomised control experiment, where *T* is randomly assigned [60]. Looking at Figure 18a, we can easily conclude that the equality is not correct since there is an open backdoor path between *T* and *Cm* ($T \leftarrow C \rightarrow Cm$). One way to adjust for a confounder and have identifiable causal effects is to block the backdoor paths. In such a case we have the following equality:

$P(Cm = cm|do(t)) = \sum_c P(cm|t,c)P(c)$, which gives an unbiased estimate for the causal effect of *T* on *Cm*. In other words, when the confounders are observed, their confounding effect is neutralised. However, when they are unobserved, a solution is to use the do operator that acts similarly with a randomised experiment, making the treatment independent from its causes as shown in Figure 20b.

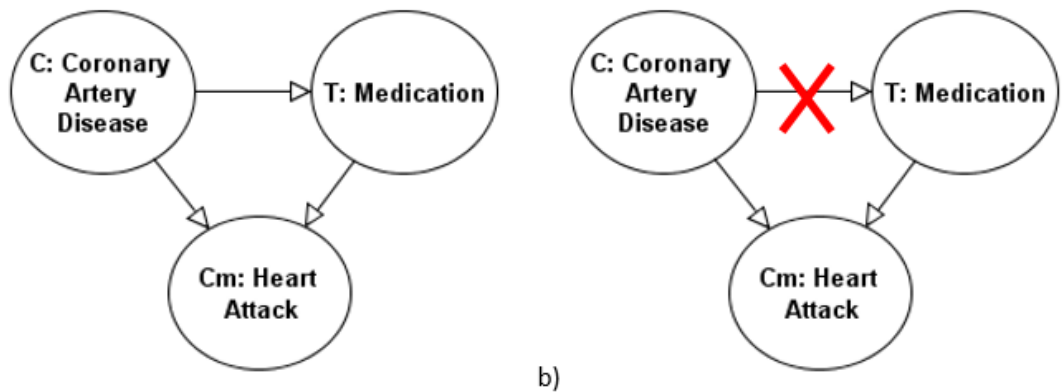

*Figure 20. (a) An example of the treatment idiom with decision arcs. (b) An example of the treatment idiom with no decision arcs*

## 4.9 Treatment reliability idiom

The treatment reliability idiom models reliability of the treatment's application. When a treatment such as surgery is performed by clinicians, the treatment's application presents as clinical practice. When a treatment such as medication is taken by the patient, the treatment's application represents how well the patient follows clinical instructions. The latter can include keeping the medication refrigerated prior to administration, taking the medication on an empty stomach, or even following frequency and dosage as prescribed by the clinician. An example of the treatment reliability idiom is shown in Figure 21. Figure 22 shows that whether the treatment is reliable or not has no impact on heart attack when the treatment is not applied.

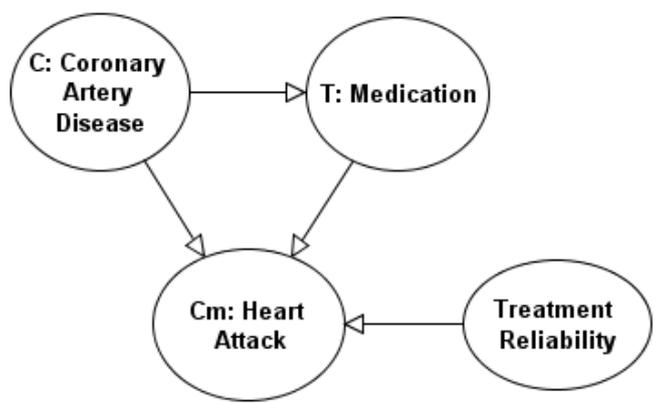

*Figure 21. An example of the treatment reliability idiom*

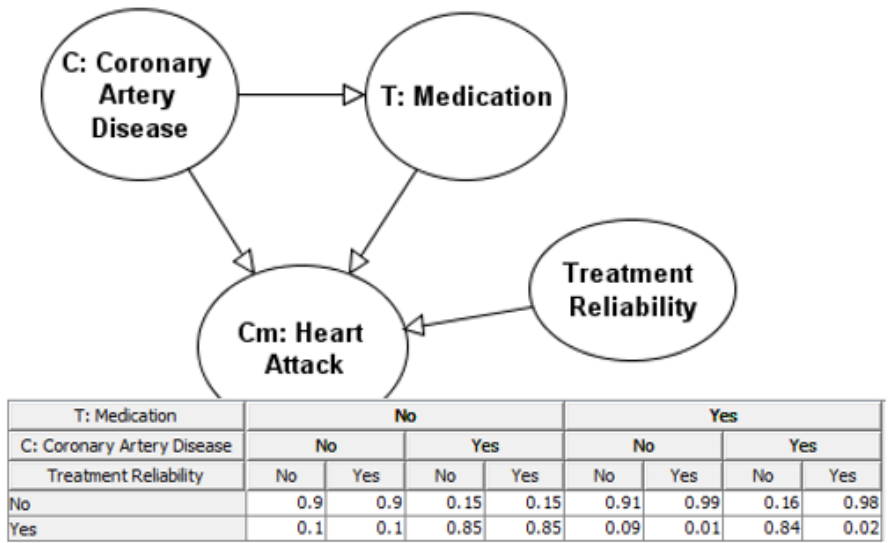

| T: Medication | No | | | | Yes | | | |
|---|---|---|---|---|---|---|---|---|
| C: Coronary Artery Disease | No | | Yes | | No | | Yes | |
| Treatment Reliability | No | Yes | No | Yes | No | Yes | No | Yes |
| No | 0.9 | 0.9 | 0.15 | 0.15 | 0.91 | 0.99 | 0.16 | 0.98 |
| Yes | 0.1 | 0.1 | 0.85 | 0.85 | 0.09 | 0.01 | 0.84 | 0.02 |

*Figure 22. An example of the treatment reliability idiom with conditional probabilities*

In Figure 23 the reasoning process is presented. At first, we observe that the patient suffers from coronary artery disease and medication is prescribed (Figure 23a). The positive treatment effect becomes more prominent when the treatment is reliable (Figure 23b). Alternately, when the treatment is not reliable, its positive effect markedly diminishes (Figure 23c).

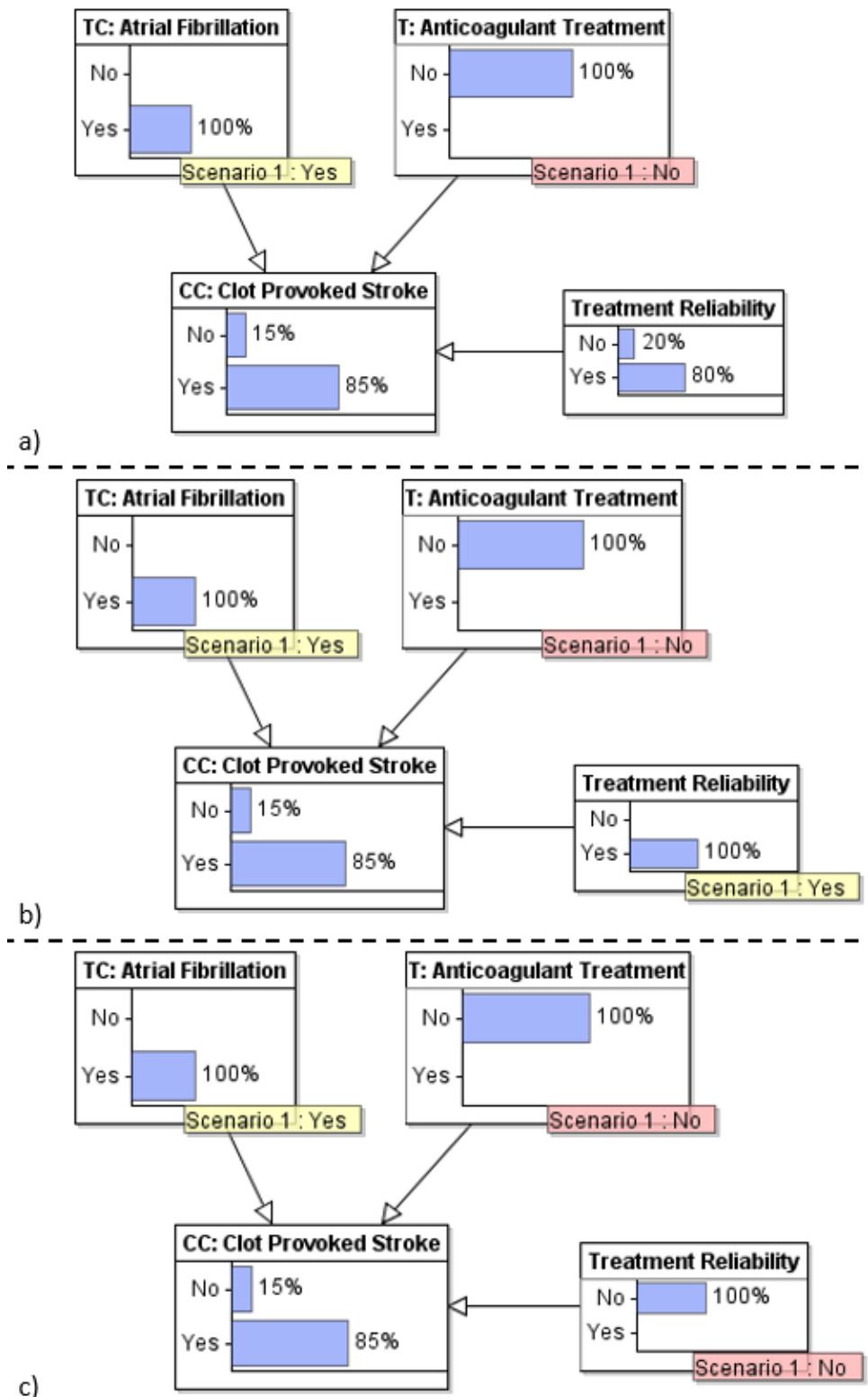

*Figure 23. (a) An example of the treatment reliability idiom, where the condition is present, and the treatment is applied. (b) An example of the treatment reliability idiom, where the condition is present, the treatment is applied, and it is reliable. (c) An example of the treatment reliability idiom, where the condition is present, the treatment is applied, and it is unreliable*

### 4.10 Counterfactual treatment idiom

Thus far we have presented idioms where reasoning about evidence or interventions is performed. However, in medical problems realistic counterfactual questions regarding the uncertain effect of hypothetical treatments may also be considered. The counterfactual treatment idiom helps us to compare what has happened with what could have happened if an alternative treatment had been applied. Using a twin network, we can compare the observed effect of an observed treatment in the actual world, with the hypothetical treatment effect in the hypothetical world. An important difference between the twin network proposed by Pearl and the counterfactual treatment idiom is that the variable on which we intervene is a treatment variable. As explained in Section 4.9, when evaluating an unconfounded treatment effect we should *adjust* for all confounders by making the treatment independent from its causes. Thus, even if Pearl suggests graph surgery only in the hypothetical world, counterfactual treatment idiom is a special case as the decision arrows that point towards the treatment should be removed not only in the hypothetical, but also in the real world. An illustrative example of the counterfactual treatment idiom is presented in Figure 24. In this example, we want to know whether heart attack would have been prevented if a proper medication had been given.

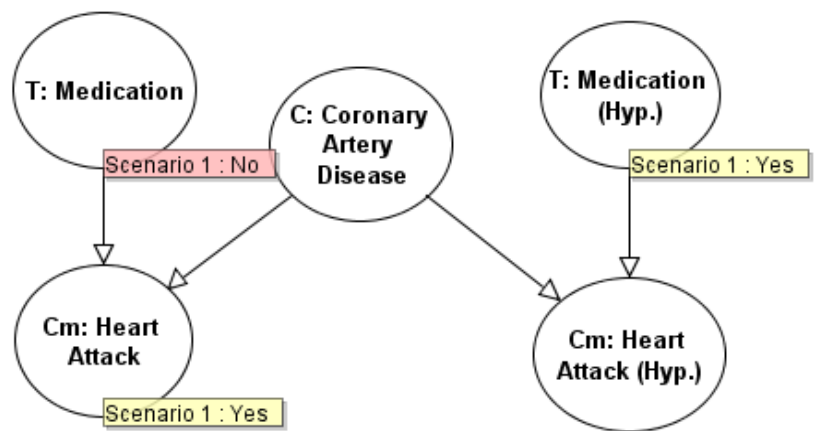

*Figure 24. An example of the counterfactual treatment idiom*

4.11   Combining the medical idioms: the core BN model for coronary artery disease

An important advantage of medical idioms is that they propose logical reasoning patterns that can be combined, reused and applied generically to help develop full medical BNs. Following the examples described above, a simplified diagnostic BN for coronary artery disease developed using medical idioms is presented in Figure 25. Based on this presented diagnostic BN, we conclude the following:

1. Not all the medical idioms may be required in a particular BN structure.

2. Any medical idiom may be used more than once.

3. It is common for medical idioms to overlap. The reasons for these overlaps are: (a) an idiom might be the subset of another idiom, such as the risk factor idiom can be the subset of the pathogenesis idiom as shown in Figure 25; and (b) a variable may have more than one role in clinical reasoning. For instance, a long bone fracture is a sign of a limb injury but at the same time it is also the risk factor for increased tissue perfusion.

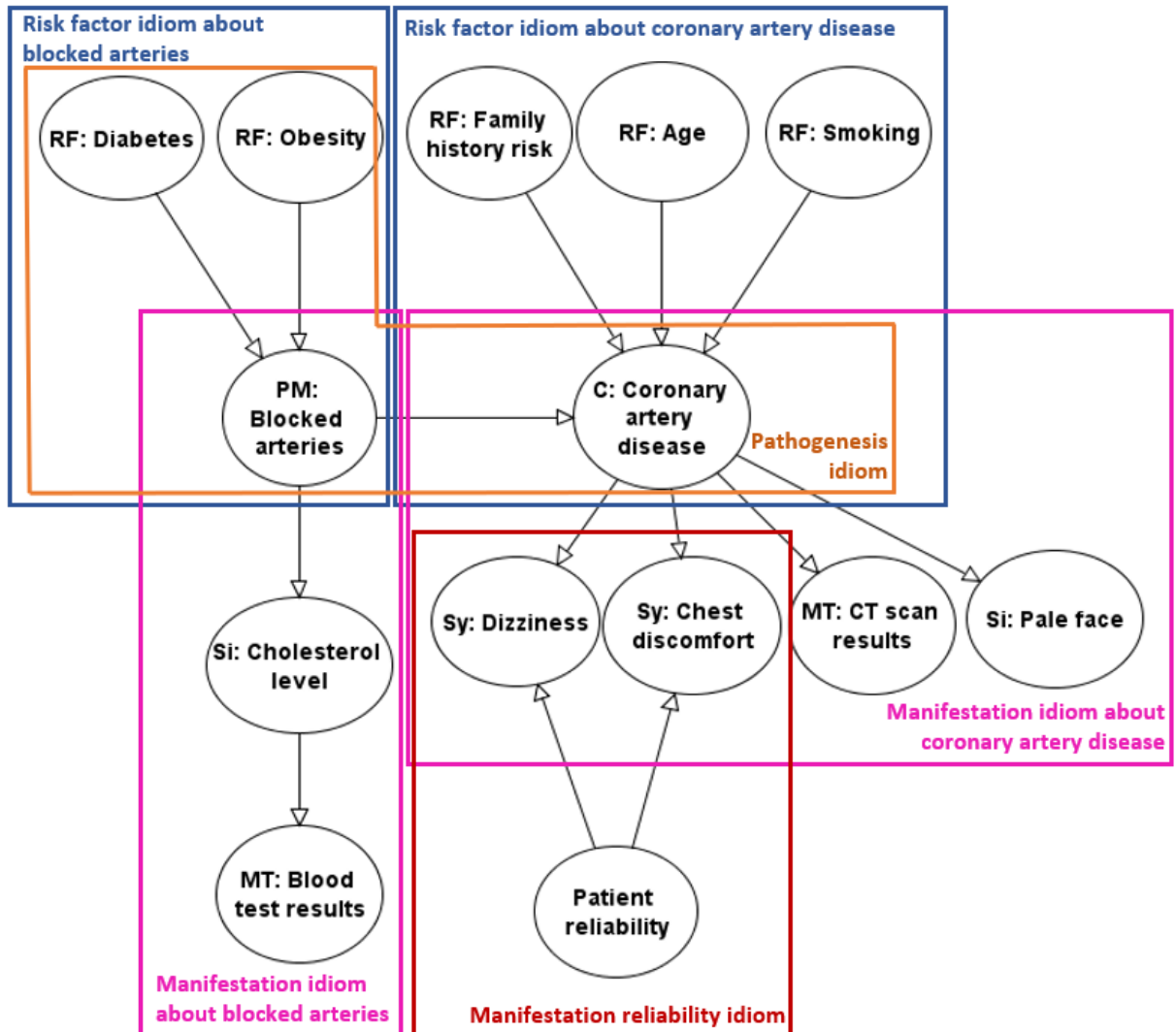

*Figure 25. Simplified diagnostic BN for cardiac problem with visible medical idioms*

## 5. Medical idiom validation

The proposed medical idioms have been inspired by a range of medical BNs developed by members of our research team over the last seven years [61], [62], [63], [64]. Currently, these medical idioms are being applied in development of medical BNs. In this section, the list of medical idioms is going to be assessed against established medical BNs.

### 5.1 Case study: assess the causal coherence of the BN structure

The example shown in Figure 26 has been published by Constantinou et al., 2015 [63]. Figure 26a illustrates a standard statistical regression model without an underlying causal structure. Figured 26b presents a claimed causal BN for head injury learned purely from data [65]. Using the proposed medical idioms, the model expert can easily understand that even if the BN structure follows the associations present in the data, it is not causally coherent. For instance, *brain scan result* is a medical test. Following the manifestation idiom, a medical test is a consequence of the outcome and not the cause of it. In addition, *delay in arrival* is a risk factor of the outcome and not a consequence of it. As a result, medical idioms can help clinicians and model experts to assess and correct the BN structure.

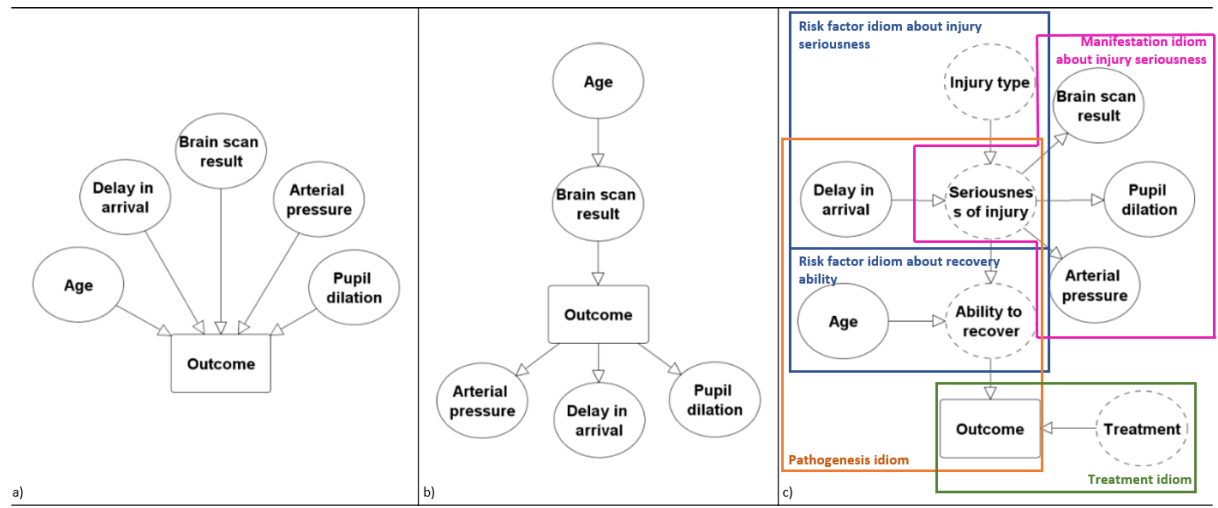

*Figure 26. (a) Standard statistical regression model learned from data. (b) "Causal" model learned purely from data proposed by Sakellaropoulos & Nikiforidis, 1999* [65]*. (c) Sensible causal model with missing/ unobserved variables proposed by Constantinou et al., 2015* [63] *embedded with medical idioms*

In Figure 26c, Costantinou et al. proposed a sensible causal structure. The proposed causal structure is in accordance with the developed medical idioms, as it is highlighted in Figure 26c. This is an additional reassurance that the causal BN structure is indeed sensible. Constantinou et al. do not explicitly describe the use of medical idioms for developing the BN structure. However, the process of using logical causal reasoning patterns has been followed unintentionally. Having a sensible causal structure may not appear to deliver an appreciable benefit when simply performing prediction or diagnosis. However, a BN with a coherent causal structure is not a black-box model and its reasoning must be explained if it is to be trusted [66], [67]. In addition, having a BN structure that represents the true causal mechanism of a disease is absolutely necessary when modelling interventions or counterfactual scenarios.

5.2 Case study: validate against a known accurate knowledge-driven medical BN

The medical BN shown in Figure 27 was developed by Yet et al. [62] to predict acute traumatic coagulopathy, with the information available, within the first 10 minutes of patient care. The BN structure was developed and refined systematically with domain experts [68]. The model has been extensively validated and shown to have a good predictive performance [62]. Figure 26 shows those medical idioms that are sufficient to connect the variables elicited from experts and are consistent with the initial causally coherent BN structure. More specifically, two main causal factors for trauma induced coagulopathy are *tissue injury* and *tissue perfusion*. Both are pathogenic mechanisms that are part of trauma induced coagulopathy's aetiology. The risk factors for *tissue injury* are the *mechanism* and *energy* of injury, as well as *injury severity* of each body part. The *injury severity* is affected again by the *mechanism* and *energy* of injury and it is manifested in the appearance of some recorded signs, symptoms and tests. For instance, an unstable pelvis is a sign of pelvic injury. All blood related manifestations for the injured body parts can cause changes in *tissue perfusion* at the same time. Pathophysiological factors such as *lactate* or *base excess* are indicators of *tissue perfusion*. Finally, dilution of blood constituents through administration of excessive prehospital resuscitation clear fluids is a causal factor for coagulopathy. In-depth medical details on the variables and causal relationships captured by this model are available in [69].

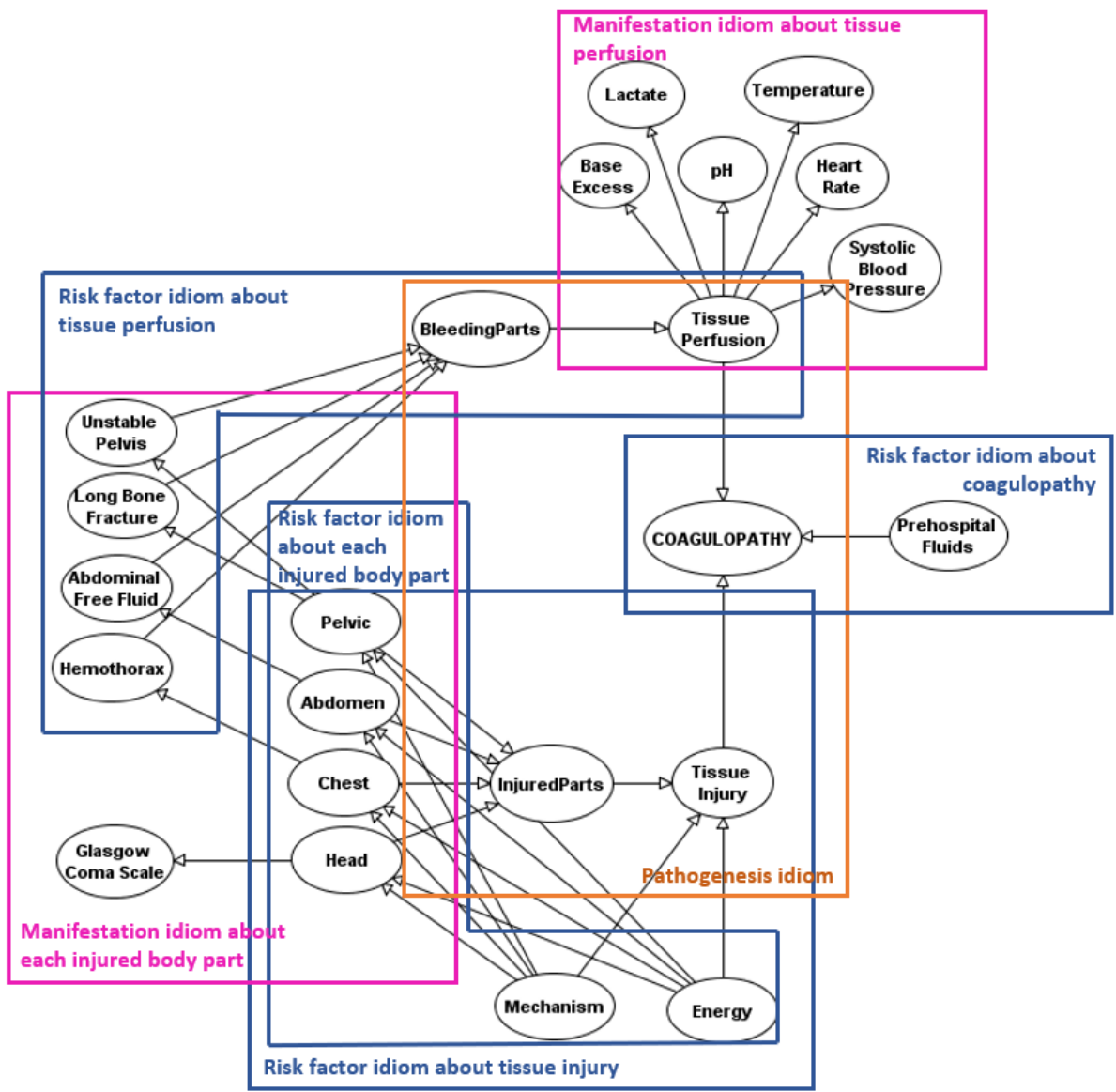

*Figure 27. Coagulopathy model, developed by Yet et al. [62], with visible medical idioms*

## 6. Conclusion

One of the major issues to consider when developing a BN using knowledge elicitation methods is to ensure a coherent structure. Most published medical BNs are presented as complete pieces of work lacking explanation of how the network's structure was developed and justification of why it represents the correct structure for the given medical application. In this paper this issue has been addressed by proposing generally applicable and reusable medical reasoning patterns, called medical idioms, that can standardise and assist in developing sensibly structured medical BNs. The proposed method complements and extends the idiom-based approach introduced by Neil, Fenton, and Nielsen in 2000. This paper presented examples of their generic idioms instantiated for specific medical purposes. Further, this paper has extended the use of idioms to represent the additional approaches of interventional and counterfactual reasoning.

The proposed medical idioms were developed as part of the EPSRC-funded PAMBAYESIAN project (https://pambayesian.org/) and are applied on two medical BNs presently under development for the conditions: (1) gestational diabetes mellitus; and, (2) rheumatoid arthritis. In this paper a third simplified case study on coronary artery disease was used to illustrate each medical idiom and the approach to combining idioms into a medical BN. The proposed idioms were assessed against medical BNs with non-sensible and sensible causal structure.

Idioms have a natural logic that mimics human reasoning processes. For instance, we would more naturally say that infection may be the cause of fever and not the alternative. Natural causal reasoning patterns can thus be represented using idioms. As described in Section 4, our proposed medical idioms act as a collection of meaningful reasoning patterns that represent essential reasoning steps in patient condition, diagnosis, prediction and management. While the medical idioms presented in this paper may not be exhaustive, they are currently capable of representing the main activities of clinical care and are sufficiently generic to be applicable in most common medical situations. Moreover, the use of medical idioms can be beneficial in the following ways:

1. **Standardise and assist medical BN development**: As demonstrated in Section 4.11 combining several medical idioms incrementally makes easier to develop cohesive medical BNs. The model expert only needs to connect elicited knowledge with the proposed idioms and reuse the most appropriate one.
2. **Assess BN structure:** As explained in Section 5, a BN structure that does not obey the idiomatic reasoning patterns is a strong indicator of a poor structure and understanding. Thus, medical idioms can help clinicians and model experts to assess and correct the BN structure, as well as improve their understanding of the underlying processes.

3. **Enhance explainability**: Based on lessons learned during the PAMBAYSIAN project, the visual representation of the medical idioms makes the BN structure easier to explain by model experts but also more easily understood by clinicians.

4. **Improve communication between model and domain experts**: Again, based on the knowledge gained during the PAMBAYESIAN project, medical idioms had a two-fold positive impact on the communication between model experts and clinicians. First, apart from helping connect the elicited variables, they were useful when eliciting experts' knowledge. Looking for the elements captured in medical idioms can guide the knowledge elicitation process. Second, and arising as a direct consequence of the previous benefit, the representation of the medical idioms in the BN structure significantly improved the review process. For model experts it was easier to explain the structure to clinicians, and for clinicians it was easier to understand and review the BN structure.

We believe that the medical idioms proposed in this paper are meaningful reasoning patterns that show the way towards a more systematic and coherent process for constructing complex medical BNs. Future work should focus on validating the proposed medical idioms against other related approaches, explained in Section 3, and exploring further the benefits described above when using medical idioms. These could bring us closer to a more generic and effective BN development process.

## Acknowledgement

All the authors acknowledge support from the Engineering and Physical Sciences Research Council (EPSRC) under project EP/P009964/1: PAMBAYESIAN: Patient Managed decision-support using Bayes Networks. We would also like to thank Dr Kudakwashe Dube for his helpful comments.